\documentclass[11pt,a4paper]{article}
\usepackage[hyperref]{acl2021}
\usepackage[export]{adjustbox}
\usepackage{amsmath}
\usepackage{amssymb}
\usepackage{amsthm}
\usepackage{appendix}
\usepackage{cancel}
\usepackage{enumitem}
\usepackage{graphicx}
\usepackage{latexsym}
\usepackage{mathtools}
\usepackage{physics}
\usepackage{stmaryrd}
\usepackage{subfiles}
\usepackage{tabularx}
\usepackage{times}

\newcommand{\lrbracket}[1]{\llbracket #1 \rrbracket}
\newcommand{\kBA}{k_{\textsf{BA}}}
\newcommand{\kE}{k_{\textsf{E}}}
\newcommand{\kEone}{k_{\textsf{E1}}}
\newcommand{\kEtwo}{k_{\textsf{E2}}}
\newcommand{\khyp}{k_{\textsf{hyp}}}
\newcommand{\khypone}{k_{\textsf{hyp1}}}
\newcommand{\khyptwo}{k_{\textsf{hyp2}}}

\newtheorem{theorem}{Theorem}
\newtheorem{corollary}{Corollary}
\newtheorem{definition}{Definition}
\allowdisplaybreaks
\usepackage{microtype}

\aclfinalcopy
\title{Conversational Negation using Worldly Context in Compositional Distributional Semantics}

\author{Benjamin Rodatz \\
  Computer Science,\\ 
  University of Oxford\\
  \texttt{benjamin.rodatz}\\
  \texttt{@cs.ox.ac.uk} \\\And
  Razin A. Shaikh \\
  Mathematical Institute,\\
  University of Oxford\\
  \texttt{razin.shaikh}\\
  \texttt{@maths.ox.ac.uk} \\
  {\centering All authors have contributed equally.}
  \And
  Lia Yeh \\
  Quantum Group, \\
  Computer Science,\\ 
  University of Oxford\\
  \texttt{lia.yeh}
  \texttt{@cs.ox.ac.uk}\\
  }

\date{6 April 2021}

\begin{document}
\maketitle
\begin{abstract}
We propose a framework to model an operational conversational negation by applying \emph{worldly context} (prior knowledge) to logical negation in compositional distributional semantics.
Given a word, our framework can create its negation that is similar to how humans perceive negation.
The framework corrects logical negation to weight meanings closer in the entailment hierarchy more than meanings further apart.
The proposed framework is flexible to accommodate different choices of logical negations, compositions, and worldly context generation. In particular, we propose and motivate a new logical negation using matrix inverse.

We validate the sensibility of our conversational negation framework by performing experiments, leveraging density matrices to encode graded entailment information.
We conclude that the combination of subtraction negation ($\neg_{sub}$) and phaser in the basis of the negated word yields the highest Pearson correlation of 0.635 with human ratings.

\end{abstract}

\section{Introduction}
Negation is fundamental to every human language, marking a key difference from how other animals communicate \cite{horn:1972negation}. It enables us to express denial, contradiction, and other uniquely human aspects of language.
As humans, we know that negation has an operational interpretation: if we know the meaning of \emph{A}, we can infer the meaning of \emph{not~A}, without needing to see or hear \emph{not~A} explicitly in any context.

Formalizing an operational description of how humans interpret negation in natural language is a challenge of significance to the fields of linguistics, epistemology, and psychology.
\citet{kruszewski:2016conv_neg} notes that there is no straightforward negation operation that, when applied to the distributional semantics vector of a word, derives a negation of that word that captures our intuition.
This work proposes and experimentally validates an operational framework for conversational negation in compositional distributional semantics.

In the field of distributional semantics, there have been developments in capturing the purely logical form of negation.
\citet{widdows:2003word} introduce the idea of computing negation by mapping a vector to its orthogonal subspace; \citet{lewis:2020towardslogicalnegation} analogously model their logical negation for density matrices.
However, logical negation alone is insufficient in expressing the nuances of negation in human language. Consider the sentences:
\begin{enumerate}[label=\alph*), topsep=0.1em, itemsep=-0.3em]
    \item \texttt{This is not an apple;\\ this is an orange.}
    \item \texttt{This is not an apple;\\ this is a paper.}
\end{enumerate}
Sentence a) is more plausible in real life than sentence b). However, since apples and oranges share a lot in common, their vector or density matrix encodings would most likely not be orthogonal. Consequently, such a logical negation of apple would more likely indicate a paper than an orange.

\citet{hermann:2013notnotbad} propose that the encoding of a word should have a distinct ``domain'' and ``value'', and its negation should only affect the ``value''. In this way, \emph{not~blue} would still be in the domain of \emph{color}. However, they do not provide any scalable way to generate such representation of ``domain'' and ``value'' from a corpus. We argue that this domain need not be encoded in the vector or density matrix itself. Instead, we propose a method to generate what we call \emph{worldly context} directly from the word and its relationships to other words, computed a priori using worldly knowledge.

Furthermore, we want such conversational negation to generalize from words to sentences and to entire texts. DisCoCat~\cite{coecke:2010DMM} provides a method to compose the meaning of words to get the meaning of sentences and DisCoCirc~\cite{coecke:2020textstructure} extends this to propagate knowledge throughout the text. Therefore, we propose our conversational negation in the DisCoCirc formalism, putting our framework in a rich expanse of grammatical types and sentence structures. Focusing on the conversational negation of single words, we leave the interaction of conversational negation with grammatical structures for future work.

Section~\ref{sec:background} introduces the necessary background. Section~\ref{sec:logical_negation} discusses the logical negation using subtraction from the identity matrix from \citet{lewis:2020towardslogicalnegation}, and proposes and justifies a second, new form of logical negation using matrix inverse. Section~\ref{sec:context_determination} introduces methods for context creation based on worldly knowledge. Section~\ref{sec:conversationalNegation} presents the general framework for performing conversational negation of a word by combining logical negation with worldly context. Section~\ref{sec:experiments} experimentally verifies the proposed framework, comparing each combination of different logical negations, compositions, bases, and worldly context generation. We end our discussion with an overview of future work.

\section{Background} \label{sec:background}
\subsection{Conversational negation}\label{sec:bg_conv_neg}

\citet{kruszewski:2016conv_neg} point out a long tradition in formal semantics, pragmatics and psycholinguistics which has argued that negation---in human conversation---is not simply a denial of information; it also indicates the truth of an \emph{alternative} assertion. They call this alternative-licensing view of negation \emph{conversational negation}.

Another view on negation states that the effect of negation is merely one of information denial \cite{evans:1996role}. However, \citet{prado:2006negation} explain that even under this view, the search for alternatives could happen as a secondary effort for interpreting negation in the sentence.

The likelihood of different alternatives to a negated word inherently admits a grading \cite{oaksford:2002contrast, kruszewski:2016conv_neg}. For example, something that is not a \emph{car} is more likely to be a \emph{bus} than a \emph{pen}. They argue that the most plausible alternatives are the ones that are applicable across many varied contexts; \emph{car} can be replaced by \emph{bus} in many contexts, but it requires an unusual context to sensibly replace \emph{car} with \emph{pen}.

\subsection{Compositional semantics and DisCoCirc}\label{sec:discocirc}
Language comprehension depends on understanding the meaning of words as well as understanding how the words interact with each other in a sentence. While the former is an understanding of the definitions of words, the latter requires an understanding of grammar. \citet{coecke:2010DMM} build on this intuition to propose DisCoCat, a compositional distributional model of meaning, making use of the diagrammatic calculus originally introduced for quantum computing \cite{Abramsky:2004categorical_semantics}. In \citet{coecke:2020textstructure}, this model was extended to DisCoCirc which generalized DisCoCat from modeling individual sentences to entire texts. In DisCoCirc, the two sentences
\begin{center}
    \texttt{Alice is an elf.}\\
    \texttt{Alice is old.}
\end{center}
are viewed as two processes updating the state of Alice, about whom, at the beginning of the text, the reader knows nothing. 
\begin{figure}
    \centering
    \includegraphics[scale=1]{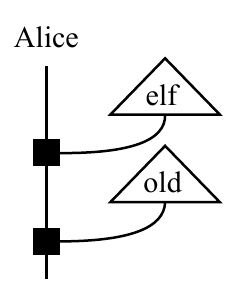}
    \caption{Graphical representation of meaning updating in DisCoCirc - read from top to bottom}
    \label{fig:discocirc_example}
\end{figure} 
Graphically this can be displayed as shown in Figure~\ref{fig:discocirc_example}. The wire labeled by \textit{Alice} represents the knowledge we have about Alice at any point in time. It is first updated by the fact that she is an elf and subsequently updated by the fact that she is old. We use a black square to represent a general meaning-update operation, which can be one of a variety of operators we discuss in the next section. DisCoCirc allows for more grammatically complex sentence and text structures not investigated in this work.

DisCoCirc allows for various ways of representing meaning such as vector spaces \cite{coecke:2010DMM, grefenstette:2011experimental_support}, conceptual spaces \cite{Bolt:2017conceptual_spaces}, and density matrices \cite{balkir:2016entailment_using_density_matrices, lewis:2019compositional_hyponymy}. A density matrix is a complex matrix, which is equal to its own conjugate transpose (Hermitian) and has non-negative eigenvalues (positive semidefinite). They can be viewed as an extension of vector spaces to allow for encoding lexical entailment structure (see Section~\ref{section:hyponymies}), a property for which they were selected as the model of meaning for this paper. 

\subsection{Compositions for meaning update}\label{sec:composition}
We present four compositions for meaning update:
\noindent\hrulefill
\begin{flalign}\label{eq:spider}
{\hspace{0.15em}\includegraphics[width=1.2em, valign=c]{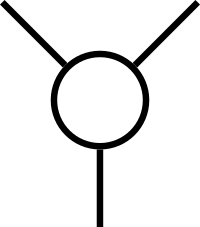} \hspace{0.15em} \emph{spider}\textsf{(A, B)}}
    \coloneqq U_s (\textsf{A} \otimes \textsf{B}) U_s^{\dagger}&&
\end{flalign}
\vspace{-2.5em}
\begin{itemize}[leftmargin=*, label={-}]
\setlength{\itemsep}{1pt}
        \setlength{\parskip}{0pt}
        \setlength{\parsep}{0pt}
\item $U_s = \sum_i \ket{i}\bra{ii}$ where $\{\ket{i}\}_i$ is \textsf{B}'s eigenbasis
\item non-linear AND in \citet{coecke:2020textstructure}
\end{itemize}
\vspace{-0.8em}
\noindent\hrulefill
\begin{flalign}\label{eq:fuzz}
{\includegraphics[width=1.2em, valign=c]{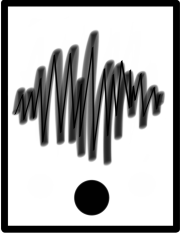}}\hspace{0.27em} \emph{fuzz}\textsf{(A, B)}
    \coloneqq\sum_i x_i P_i \circ \textsf{A} \circ P_i&&
\end{flalign}
\vspace{-2.5em}
\begin{itemize}[leftmargin=*, label={-}]
\setlength{\itemsep}{1pt}
        \setlength{\parskip}{0pt}
        \setlength{\parsep}{0pt}
\item $\textsf{B} = \sum_i x_i P_i$
\item in \citet{coecke:2020meaningupdate}
\item Kmult in \citet{lewis:2020towardslogicalnegation}
\end{itemize}
\vspace{-0.8em}
\noindent\hrulefill
\begin{flalign}\label{eq:phaser}
{\includegraphics[width=1.2em, valign=c]{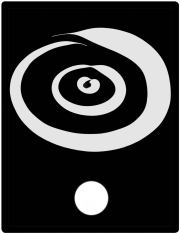}}\hspace{0.27em} \emph{phaser}\textsf{(A, B)} \coloneqq \textsf{B}^{\frac{1}{2}}\textsf{A}\textsf{B}^{\frac{1}{2}}&&
\end{flalign}
\vspace{-2.5em}
\begin{itemize}[leftmargin=*, label={-}]
\setlength{\itemsep}{1pt}
        \setlength{\parskip}{0pt}
        \setlength{\parsep}{0pt}
\item $\textsf{B} = \sum_i x_i^2 P_i \text{ where } \textsf{B}^{\frac{1}{2}} = \sum_i x_i P_i$
\item in \citet{coecke:2020meaningupdate}
\item Bmult in \citet{lewis:2020towardslogicalnegation}
\item corresponds to quantum Bayesian update \cite{vandewetering:2018ordering}
\end{itemize}
\vspace{-0.8em}
\noindent\hrulefill
\begin{flalign}\label{eq:diag}
{\large \scalebox{-1}[1]{\cancel{$\square$}}}\hspace{0.4em} \emph{diag}\textsf{(A, B)}
        \coloneqq\emph{dg}(\textsf{A}) \circ \emph{dg}(\textsf{B})&&
\end{flalign}
\vspace{-2.5em}
\begin{itemize}[leftmargin=*, label={-}]
\setlength{\itemsep}{1pt}
        \setlength{\parskip}{0pt}
        \setlength{\parsep}{0pt}
\item a Compr from \citet{delascuevas:2020catsclimb}: lifts verbs and adjectives to completely positive maps matching their grammatical type
\end{itemize}

\noindent where \textsf{A} and \textsf{B} are density matrices, $x_i$ is a real scalar between 0 and 1, $P_i$'s are projectors, and the function $\emph{dg}$ sets all off-diagonal matrix elements to 0 giving a diagonal matrix.

Of the many Compr variants \cite{delascuevas:2020catsclimb}, we only consider \emph{diag} and \emph{mult} (elementwise matrix multiplication, which is an instance of \emph{spider}) as candidates for composition. All other variants are scalar multiples of one input, the identity wire, or a maximally mixed state; therefore we do not consider them as they discard too much information about the inputs.

For \emph{spider}, \emph{fuzz}, and \emph{phaser}, choosing the basis of the composition determines the basis the resulting density matrix takes on, and its meaning is interpreted in \cite{coecke:2020meaningupdate}.

\subsection{Lexical entailment via hyponymies}
\label{section:hyponymies}
A word $w_A$ is a hyponym of $w_B$ if $w_A$ is a type of $w_B$; then, $w_B$ is a hypernym of $w_A$. For example, \emph{dog} is a hyponym of \emph{animal}, and \emph{animal} is a hypernym of \emph{dog}.
Where there is a meaning relation between two words, there exists an entailment relation between two sentences containing those words.
Measures to quantify these relations ought to be \textit{graded}, as one would expect some entailment relations to be weaker than others. Furthermore, such measures should be \textit{asymmetric} (a bee is an insect, but an insect is not necessarily a bee) and \textit{pseudo-transitive} (a t-shirt is a shirt, a shirt can be formal, but a t-shirt is usually not formal).

One of the limitations of the vector space model of NLP is that it does not admit a natural non-trivial graded entailment structure \cite{balkir:2016entailment_using_density_matrices, coecke:2020textstructure}.
\citet{bankova:2019graded_entailment} utilize the richer setting of density matrices to define a measure called $k$-hyponymy, generalizing the L\"owner order to have a grading for positive operators, satisfying the above three properties. They further lift from entailment between words to between two sentences of the same grammatical structure, using compositional semantics, and prove a lower bound on this entailment between sentences.

The $k$-hyponymy ($\khyp$) between density matrices \textsf{A} and \textsf{B} is the maximum $k$ such that
\begin{equation}
\label{eq:khyp}
    \textsf{A} \sqsubseteq_k \textsf{B} \Longleftrightarrow \textsf{B} - k\,\textsf{A} \text{  is a positive operator}
\end{equation}
where $k$ is between 0 (no entailment) and 1 (full entailment).

\Citet{vandewetering:2018ordering} finds that the crisp L\"owner ordering ($\khyp = 1$) is trivial when operators are normalized to trace 1. On the other hand, they enumerate highly desirable properties of the L\"owner order when normalized to highest eigenvalue 1. In particular, the maximally mixed state is the bottom element; all pure states are maximal; and the ordering is preserved under any linear trace-preserving isometry (including unitaries), convex mixture, and the tensor product. In our experiments, we leverage these ordering properties following \citet{lewis:2020towardslogicalnegation}'s convention of normalizing operators to highest eigenvalue $\leq 1$.

According to \citet[Theorem 2]{bankova:2019graded_entailment}, when $supp(\textsf{A}) \subseteq supp(\textsf{B})$, $\khyp$ is given by $1/\gamma$, where $\gamma$ is the maximum eigenvalue of $\textsf{B}^+\textsf{A}$. Here $\textsf{B}^+$ denotes the Moore-Penrose inverse of $\textsf{B}$, which we refer to in the next section as support inverse.
If $supp(\textsf{A}) \not\subseteq supp(\textsf{B})$, $\khyp$ is 0.
This means that $\khyp$ admits a grading, but is not robust to errors.
In our experiments, to circumvent this issue of almost all of our calculated $\khyp$ being $0$, we employ a generalized form of $\khyp$ equivalent to as originally defined in \citet[Theorem 2]{bankova:2019graded_entailment}, less checking whether $supp(\textsf{A}) \subseteq supp(\textsf{B})$.

To propose more robust measures, \citet{lewis:2019compositional_hyponymy} says \textsf{A} entails \textsf{B} with the error term \textsf{E} if there exists a \textsf{D} such that:
\begin{equation}
    \label{equation:error}
    \textsf{A} + \textsf{D} = \textsf{B} + \textsf{E}
\end{equation}
to define the following two entailment measures
\begin{equation}
    \label{equation:kBA}
    \kBA = \frac{\sum_i \lambda_i}{\sum_i \abs{\lambda_i}} = \frac{\textsf{Trace}(\textsf{D} - \textsf{E})}{\textsf{Trace}(\textsf{D} + \textsf{E})}
\end{equation}
\begin{equation}
    \label{equation:kE}
    \kE = 1 - \frac{\norm{\textsf{E}}}{\norm{\textsf{A}}}
\end{equation}
where the $\lambda_i$'s are the eigenvalues of \textsf{B} $-$ \textsf{A}.
In Equations \ref{equation:kBA} and \ref{equation:kE}, the error term \textsf{E} satisfying Equation \ref{equation:error} is constructed by taking the diagonalization of \textsf{B} $-$ \textsf{A}, setting all positive eigenvalues to zero, and changing the sign of all negative eigenvalues. $\kBA$ ranges from $-1$ to $1$, and $\kE$ ranges from $0$ to~$1$.

According to \citet{delascuevas:2020catsclimb}, \emph{diag}, \emph{mult}, and \emph{spider} preserve crisp L\"owner order:
\begin{equation}
    \label{eq:preservek}
    \textsf{A}_1 \sqsubseteq \textsf{B}_1, \textsf{A}_2 \sqsubseteq \textsf{B}_2 \Longleftrightarrow \textsf{A}_1 \  {\includegraphics[height=1em, valign=c]{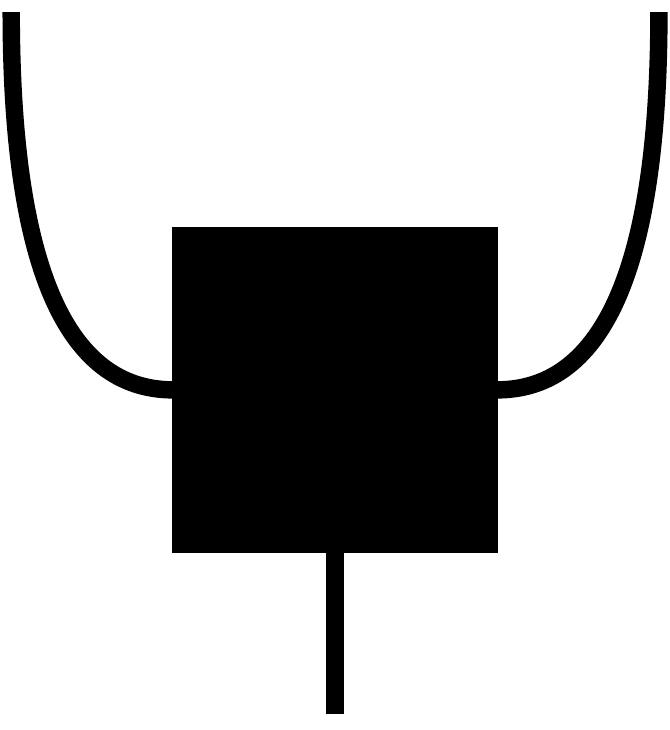}} \  \textsf{ A}_2 \sqsubseteq \textsf{B}_1 \  {\includegraphics[height=1em, valign=c]{figures/conjunction_symmetric.pdf}} \  \textsf{ B}_2
\end{equation}
\emph{Fuzz} and \emph{phaser} do not satisfy Equation~\ref{eq:preservek}.

\section{Logical negations}\label{sec:logical_negation}
To construct conversational negation, we must first define a key ingredient -- logical negation, denoted by $\neg$. The logical negation of a density matrix is a unary function that yields another density matrix.

The most important property of a logical negation is that it must interact well with hyponymy. Ideally, the interpretation of the contrapositive of an entailment must be sensible:
\begin{equation}
\textsf{A} \sqsubseteq \textsf{B} \Longleftrightarrow \neg \textsf{B} \sqsubseteq \neg \textsf{A}
\label{equation:contrapositive}
\end{equation}

A weakened notion arises from allowing varying degrees of entailment:
\begin{equation}
\textsf{A} \sqsubseteq_{k} \textsf{B} \Longleftrightarrow \neg \textsf{B} \sqsubseteq_{k'} \neg \textsf{A}
\label{equation:contrapositivek}
\end{equation}
where $k = k'$ in the ideal case.

Equation~\ref{equation:contrapositivek} necessitates any candidate of logical negation to be \textit{order-reversing}.
However, \citet{vandewetering:2018ordering} proved that all unitary operations preserve L\"owner order. Therefore, no quantum gates can reverse L\"owner order, and the search for a logical negation compatible with quantum natural language processing \cite{coecke:2020QNLP} (originally formulated in the category of \textbf{CPM}(\textbf{FHilb}) \cite{piedeleu:2015open}) remains an open question.

We now discuss two candidates for logical negation that have desirable properties and interaction with the hyponymies presented in Section~\ref{section:hyponymies}.

\subsection{Subtraction from identity negation}
\citet{lewis:2020towardslogicalnegation} introduces a candidate logical negation which preserves positivity of density matrix $\textsf{X}$:
\begin{equation}
\label{eq:id_minus}
    \neg_{sub} \textsf{X} \coloneqq \mathbb{I} - \textsf{X}
\end{equation}
In the case where \textsf{X} is a pure state, it maps \textsf{X} to the subspace orthogonal to it, as the identity matrix $\mathbb{I}$ is the sum of orthonormal projectors.
This logical negation satisfies Equation~\ref{equation:contrapositive} for the crisp L\"owner order. It satisfies Equation~\ref{equation:contrapositivek} with $k = k'$ for $\kBA$, but not for $\khyp$ or $\kE$.

\subsection{Matrix inverse negation}
We introduce a new candidate for logical negation, the \emph{matrix inverse}. This reverses L\"owner order, i.e. satisfies Equation~\ref{equation:contrapositivek} with $k = k'$ (see Corollary~\ref{cor:invkhyp}
in Appendix). It additionally satisfies Equation~\ref{equation:contrapositivek} with $k = k'$ for $\kBA$ if both density operators have same eigenbases (see Theorem~\ref{theorem:inversekBA} 
in Appendix).

As the matrix inverse of a non-invertible matrix is undefined, we define a logical negation from two generalizations of the matrix inverse acting upon the support and kernel subspaces, respectively.

\begin{definition}
\label{def:suppinv}
For any density matrix \textsf{X} with spectral decomposition \textsf{X} $= \sum_i \lambda_i \ket{i}\bra{i}$,
\begin{equation}
    \label{eq:suppinv}
    \neg_{supp} \textsf{X} \coloneqq \sum_i
    \begin{cases}
    \frac{1}{\lambda_i} \ket{i}\bra{i},& \text{if } \lambda_i > 0\\
    0,              & \text{otherwise}
\end{cases}
\end{equation}
\end{definition}
Definition~\ref{def:suppinv} is the Moore-Penrose generalized matrix inverse and is equal to the matrix inverse when the kernel is empty. It has the property that Equation~\ref{equation:contrapositivek} with $k = k'$ is satisfied for $\khyp$ when $rank(\textsf{A}) = rank(\textsf{B})$ (see Theorem~\ref{theorem:suppinvkhyp}
in Appendix). We call it the \emph{support inverse}, to contrast with what we call the \emph{kernel inverse}:
\begin{definition}
\label{def:kerinv}
For any non-invertible density matrix \textsf{X} with spectral decomposition \textsf{X} $= \sum_i \lambda_i \ket{i}\bra{i}$,
\begin{equation}
    \neg_{ker} \textsf{X} \coloneqq \sum_i
    \begin{cases}
    1 \ket{i}\bra{i},& \text{if } \lambda_i = 0\\
    0,              & \text{otherwise}
\end{cases}
\end{equation}
\end{definition}
The kernel inverse is the limit of matrix regularization by spectral filtering (i.e. setting all zero eigenvalues to an infinitesimal positive eigenvalue), then inverting the matrix and normalizing to highest eigenvalue 1. Its application discards all information about the eigenspectrum of the original matrix. Therefore, applying the kernel inverse twice results in a maximally mixed state over the support of the original matrix. Operationally speaking, $\neg_{ker}$ and $\neg_{sub}$ act upon the kernel of the original matrix identically.

We can think conceptually of a negated word as containing elements both ``near'' (in support) and ``far'' (in kernel) from the original word. Therefore, a logical negation should encompass nonzero values in the original matrix's support and in its kernel; it is up to conversational negation to then weight the values in the logical negation according to their contextual relevance.

On their own, neither the support inverse nor the kernel inverse are sensible candidates for logical negation. 
A convex mixture of the two, which we call \emph{matrix inverse} and denote with $\neg_{inv}$, spans both support and kernel of the original matrix. In our experiments we weight support and kernel equally, but other weightings could be considered, for instance to take into account a noise floor or enforce the naively unsatisfied property that twice application is the identity operation.

When composing a density matrix \textsf{X} with $\neg_{inv} \textsf{X}$ or $\neg_{supp} \textsf{X}$ via \emph{spider}, \emph{fuzz}, or \emph{phaser}, the resulting density matrix has the desired property of being a maximally mixed state on the support with zeroes on the kernel (see Theorem~\ref{theorem:compmaxmixed} 
and Corollary~\ref{cor:compmaxmixed} 
in Appendix). In other words, this operation is the fastest ``quantum (Bayesian, in the case of \emph{phaser}) update'' from a density matrix to the state encoding no information other than partitioning support and kernel subspaces. Interpreting composition as logical AND, this corresponds to the contradiction that a proposition (restricted to the support subspace) cannot simultaneously be true and not true.

\subsection{Normalization}
$\neg_{sub}$, $\neg_{supp}$, and $\neg_{inv}$ preserve eigenvectors (up to uniqueness for eigenvalues with multiplicity $> 1$). We ignore normalization for logical negation because in our conversational negation framework, which we introduce in Section~\ref{sec:conversationalNegation}, we can always normalize to largest eigenvalue $\leq 1$ after the composition operation.

\section{Context determination} \label{sec:context_determination}

Negation is intrinsically dependent on context. Context can be derived from two sources: 1) knowledge gained throughout the sentence or the text (textual context), and 2) worldly knowledge from experts or data such as a corpus (worldly context). While textual context depends on the specific text being analyzed, worldly context can be computed a priori. In this section, we introduce worldly context and propose two methods of computing it.

\subsection{Worldly context}\label{sec:worldly_context}
Worldly knowledge is a certain understanding of the world that most users of a language intuitively possess. We want to capture this worldly knowledge to provide a context for negation that is not explicit in the text. In this section, we propose two methods of generating a worldly context: 1) knowledge encoded in an entailment hierarchy such as WordNet, and 2) generalizing the ideas of the first method to context derivation from the entailment information encoded in density matrices.

\subsubsection{Context from an entailment hierarchy} \label{sec:wordnet}
\begin{figure}
    \centering
    \includegraphics[scale=1]{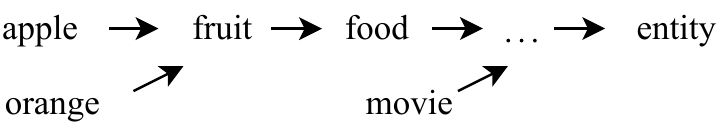}
    \caption{Example of hyponymy structure as can be found in entailment hierarchies}
    \label{fig:tree_hierarchy}
\end{figure}

We consider an entailment hierarchy for words that leads to relations such as in Figure~\ref{fig:tree_hierarchy}, where a directed edge can be understood as a hyponym relation. Such relational hierarchy can be obtained from human curated database like WordNet~\cite{wordnet} or using unsupervised methods such as Hearst patterns~\citep{hearst:1992automatic, roller:2018hearst}.

We can use such a hierarchy of hyponyms to generate worldly context, as words usually appear in the implicit context of their hypernyms; for example, \textit{apple} is usually thought of as a \textit{fruit}. Now, to calculate the worldly context for the word \textit{apple}, we take a weighted sum of the hypernyms of \textit{apple}, with more direct hypernyms such as \textit{fruit} weighted higher than more distant hypernyms such as \textit{entity}. This corresponds to the idea that when we talk in the context of \textit{apple}, we are more likely to talk about an \textit{orange} (hyponym of \textit{fruit}) than a \textit{movie} (hyponym of \textit{entity}). Hence, for a word $w$ with hypernyms $h_1, \ldots, h_n$ ordered from closest to furthest, we define the worldly context $\texttt{wc}_{w}$ as:
\begin{equation} \label{eq:wc_wordnet}
    \lrbracket{\texttt{wc}_{w}} \coloneqq \sum_i p_i \lrbracket{h_i}
\end{equation}
where $p_i \geq p_{i+1}$ for all $i$.

For this approach, we assume that the density matrix of the word is a mixture containing its hyponyms; i.e. the density matrix of \textit{fruit} is a mixture of all fruits such as \textit{apple}, \textit{orange} and \textit{pears}.

\subsubsection{Context using entailment encoded in the density matrices}\label{sec:data-driven-context}

As explained in Section~\ref{section:hyponymies}, density matrix representation of words can be used to encode the information about entailment between words. Furthermore, this entailment can be graded; for example, \textit{fruit} would entail \textit{dessert} with a high degree, but not necessarily by~1. Such graded entailment is not captured in the human curated WordNet database. Although there have been proposals to extend WordNet \cite{boyd:2006extending_wordnet, ahsaee:2014semantic}, such semantic networks are not yet available.

We generalize the idea of entailment hierarchy by considering a directed weighted graph where each node is a word and the edges indicate how much one word entails the other. Once we have the density matrices for words generated from corpus data, we can build this graph by calculating the graded hyponymies (see Section~\ref{section:hyponymies}) among the words, thereby extracting the knowledge gained from the corpus encoded in the density matrices, without requiring human narration. 

Consider words $x$ and $y$ where $x \sqsubseteq_p y$ and $y \sqsubseteq_q x$. In the ideal case, there are three possibilities: 1)~$x$ and $y$ are not related (both $p$ and $q$ are small), 2)~one is a type of the other (one of $p$ and $q$ is large), or 3)~they are very similar (both $p$ and $q$ are large). Hence, we need to consider both $p$ and $q$ when we generate the worldly context. To obtain the worldly context for a word $w$, we consider all nodes (words) connected to $w$ along with their weightings. If $p_1, \ldots, p_n$ and $q_1, \ldots, q_n$ are the weights of the edges from $w$ to words $h_1, \ldots, h_n$, then worldly context $\texttt{wc}_w$ is given by
\begin{equation} \label{eq:wc_hyp}
    \lrbracket{\texttt{wc}_w} \coloneqq \sum_i f(p_i, q_i) \lrbracket{h_i}
\end{equation}
where $f$ is some function of weights $p_i$ and $q_i$.

\section{Conversational negation in DisCoCirc}
\label{sec:conversationalNegation}

\subsection{A framework for conversational negation}
In this section, we present a framework to obtain conversational negation by composing logical negation with worldly context. As discussed in Section~\ref{sec:bg_conv_neg}, negation---when used in conversation---can be viewed as not just a complement of the original word, but as also suggesting an \emph{alternative} claim. Therefore, to obtain conversational negation, we need to adapt the logical negation to take into account the worldly context of the negated word. 

In DisCoCirc (see Section~\ref{sec:discocirc}), words are wires, and sentences are processes that update meaning of the words. Similarly, we view \emph{conversational negation} as a process that updates the meaning of the words. We propose the general framework for conversational negation by defining it to be the logical negation of the word, updated through composition with the worldly context evoked by that word:
\begin{equation}
    \includegraphics[scale=0.9, valign=c]{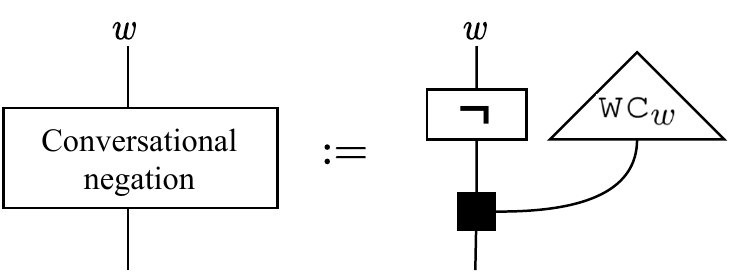}
    \label{fig:conv_neg_framework}
\end{equation}

The framework presented here is general; i.e. it does not restrict the choice of logical negation, worldly context or composition. 
The main steps of conversational negation are:
\begin{enumerate}[itemsep=-0.5em, topsep=0em]
    \item Calculate the logical negation $\neg(\lrbracket{w})$.
    \item Compute the worldly context $\lrbracket{\texttt{wc}_w}$.
    \item Update the meaning of $\neg(\lrbracket{w})$ by composing with $\lrbracket{\texttt{wc}_w}$ to obtain $\neg(\lrbracket{w})\  \includegraphics[height=1em, valign=c]{figures/conjunction_symmetric.pdf} \ \lrbracket{\texttt{wc}_w}$.
\end{enumerate}

\noindent Further meaning updates can be applied to the output of conversational negation using compositional semantics as required from the structure of the text, although we do not investigate this in the current work.

\subsection{See it in action}\label{sec:local_context}
We present a toy example to develop intuition of how meaning provided by worldly context interacts with logical negation and composition to derive conversational negation.
Suppose $\{apple,\ orange,\ fig,\ movie\}$ are pure states forming an orthonormal basis (ONB). In practice ONBs are far larger, but this example suffices to illustrate how the conversational negation accounts for which states are relevant.
We take $\neg_{sub}$ as the choice of negation and \emph{spider} in this ONB as the choice of composition.

Now, consider the sentence:
\begin{center}
    \texttt{This is not an apple.}
\end{center}
Although in reality the worldly context of $apple$ encompasses more than just $fruit$, for ease of understanding, assume the worldly context of apple is $\lrbracket{\texttt{wc}_{apple}} = \lrbracket{fruit}$, given by
\begin{align*}
    \lrbracket{fruit} = \frac{1}{2} \lrbracket{apple} + \frac{1}{3} \lrbracket{orange} + \frac{1}{6} \lrbracket{fig}
\end{align*}
Applying $\neg_{sub}(\lrbracket{apple}) = \mathbb{I} - \lrbracket{apple}$, we get
\begin{align*}
    \neg_{sub}(\lrbracket{apple}) = \lrbracket{orange} + \lrbracket{fig} + \lrbracket{movie}
\end{align*}

Finally, to obtain conversational negation, logical negation is endowed with meaning through the application of worldly context.
\begin{align*}
    \neg_{sub}(\lrbracket{apple}) {\includegraphics[height=1em]{figures/spider.png}} \lrbracket{fruit} = \frac{1}{3}\lrbracket{orange} + \frac{1}{6}\lrbracket{fig}
\end{align*}
This conversational negation example not only yields all \emph{fruits} which are not \emph{apples}, but also preserves the proportions of the non-apple fruits.

\section{Experiments}
\label{sec:experiments}
To validate the proposed framework, we perform experiments on the data set of alternative plausibility ratings created by~\citet{kruszewski:2016conv_neg}\footnote{The data set is available at \url{http://marcobaroni.org/PublicData/alternatives\_dataset.zip}}. 
In their paper, \citet{kruszewski:2016conv_neg} predict plausibility scores for word pairs consisting of a negated word and its alternative using various methods to compare the similarity of the words. While achieving a high correlation with human intuition, they do not provide an operation to model the outcome of a conversational negation. Through the experiments, we test whether our operational conversational negation still has correlation with human intuition.

\subsection{Data}
The \citet{kruszewski:2016conv_neg} data set consists of word pairs containing a noun to be negated and an alternative noun, along with a plausibility rating. We will denote the word pairs as $(w_N, w_A)$. The authors transform these word pairs into simple sentences of the form: \emph{This is not a $w_N$, it is a $w_A$} (e.g. This is not a radio, it is a dad.). These sentences are then rated by human participants on how plausible they are to appear in a natural conversation.

To build these word pairs, \citet{kruszewski:2016conv_neg} randomly picked 50 common nouns as $w_N$ and paired them with alternatives that have various relations to $w_N$.
Then using a crowd-sourcing service, they asked the human participants to judge the plausibility of each sentence. The participants were told to rate each sentence on a scale of 1 to 5.

\subsection{Methodology}
\label{subsec:experiments_methodology}
We build density matrices from 50 dimensional GloVe~\cite{pennington:2014glove} vectors using the method described in~\citet{lewis:2019compositional_hyponymy}. Then for each word pair $(w_N, w_A)$ in the data set, we use various combinations of operations to perform conversational negation on the density matrix of $w_N$ and calculate similarity with the density matrix of~$w_A$.

For conversational negation, we experiment with different combinations of logical negations, composition operations and worldly context. We use two types of logical negations: $\neg_{sub}$ and $\neg_{inv}$. For composition, we use \emph{spider}, \emph{fuzz}, \emph{phaser}, \emph{mult} and \emph{diag}. With \emph{spider}, \emph{fuzz} and \emph{phaser}, we perform experiments in two choices of basis: `w', the basis of $\neg(\lrbracket{w_N})$, and `c', the basis of $\lrbracket{\texttt{wc}_{w_N}}$. We use worldly context generated from the WordNet entailment hierarchy as per Section~\ref{sec:wordnet}; we experiment with different methods to calculate the weights $p_i$ along the hypernym path.

To find plausibility ratings, we calculate hyponymies $\khyp$, $\kE$ and $\kBA$, as well as \emph{trace similarity} (the density operator analog of cosine similarity for vectors), between the density matrix of the conversational negation of $w_N$ and $\lrbracket{w_A}$. Note that in our experiments, unlike in the originally proposed formulation of $\khyp$, we generalize $\khyp$ to not be~$0$ when $supp(\textsf{A}) \not\subseteq supp(\textsf{B}$), as described in Section \ref{section:hyponymies}.
We calculate entailment in both directions for $\kE$ and $\khyp$, which are asymmetric. The entailment from $w_N$ to $w_A$ is denoted $\kEone$ and $\khypone$ while the entailment from $w_A$ to $w_N$ is denoted $\kEtwo$ and $\khyptwo$. Finally, we calculate the Pearson correlation between our plausibility ratings and the mean human plausibility ratings from \citet{kruszewski:2016conv_neg}.

\subsection{Results}
\begin{figure}
    \centering
    \includegraphics[width=\columnwidth]{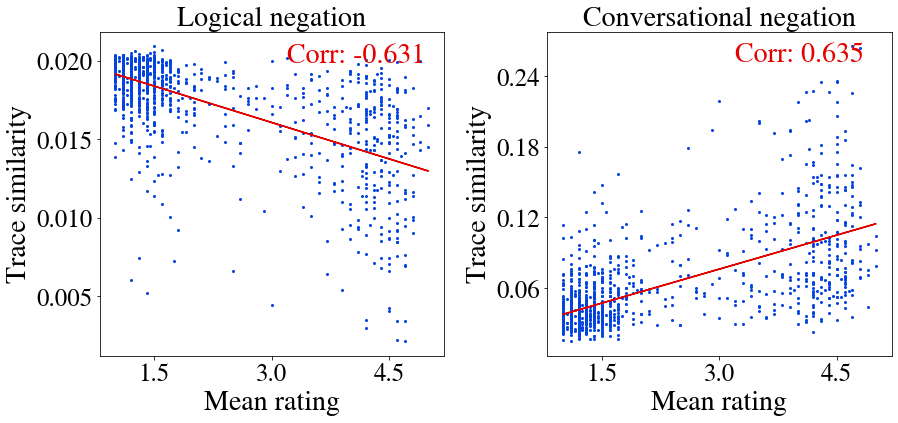}
    \caption{Correlation of logical (left) and conversational negation (right) with mean human rating}
    \label{fig:best_neg_scatter}
\end{figure}

Our experiments revealed that the best conversational negation is obtained by choosing $\neg_{sub}$ with \emph{phaser} in the basis `w'.  We achieve 0.635 correlation of the \emph{trace similarity} plausibility rating with the human ratings, as shown in Figure~\ref{fig:best_neg_scatter} (right).

On the other hand, Figure~\ref{fig:best_neg_scatter} (left) shows \emph{trace similarity} of $\neg_{sub}$ without applying any context. We observe that simply performing logical negation yields a negative correlation with human plausibility ratings. This is because logical negation gives us a density matrix furthest from the original word, going against the observation of \citet{kruszewski:2016conv_neg} that an alternative to a negated word appears in similar contexts to it. Figure~\ref{fig:best_neg_scatter} (right) shows the results of combining this logical negation with worldly context to obtain meaning that positively correlates with how humans think of negation in conversation. 

\begin{figure}
    \centering
    \includegraphics[width=\columnwidth]{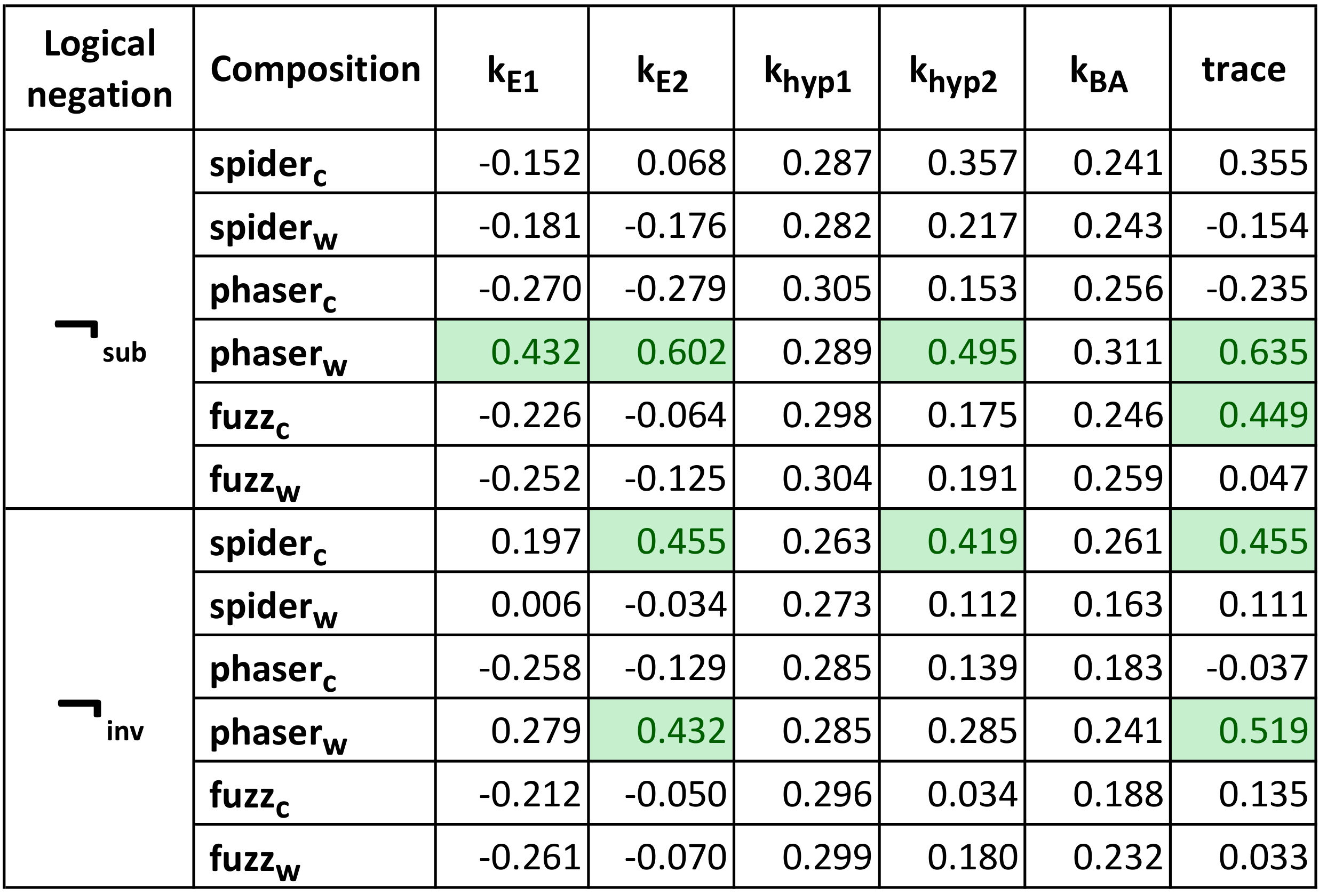}
    \caption{Correlation of various conversational negations with mean plausibility ratings of human participants. Correlations above 0.4 are highlighted in green.}
    \label{fig:results}
\end{figure}

We tested many combinations for conversational negation enumerated in Section~\ref{subsec:experiments_methodology}. The correlation between plausibility ratings for our conversational negation and the mean human plausibility rating is shown in Figure~\ref{fig:results}. We left out \emph{mult} and \emph{diag} from the table as they did not achieve any correlation above 0.3. Now, we will explore each variable of our experiments individually in the next sections.

\subsubsection{Logical negation}
We tested $\neg_{sub}$ and $\neg_{inv}$ logical negations. We found that the conversational negations built from $\neg_{sub}$ negation usually had a higher correlation with human plausibility ratings, with the highest being 0.635 as shown in Figures~\ref{fig:best_neg_scatter}~and~\ref{fig:results}. 
One exception to this is when the $\neg_{inv}$ is combined with \emph{spider} in the basis `c', for which we get the correlation of 0.455 for both \emph{trace similarity} and $\kEtwo$.

\subsubsection{Composition}
We investigated five kinds of composition operations: \emph{spider}, \emph{fuzz}, \emph{phaser}, \emph{mult}, and \emph{diag}. We found that the results using \emph{mult} and \emph{diag} do not have any statistically significant correlation ($<$0.3) with human plausibility rating. On the other hand, \emph{phaser} (in the basis `w') has the highest correlation. It performs well with both logical negations. Plausibility ratings for \emph{phaser} with $\neg_{sub}$ negation measured using $\kEtwo$ and \emph{trace similarity} has correlations of 0.602 and 0.635 respectively. \emph{Spider} and \emph{fuzz} have statistically relevant correlation for a few cases but never more than 0.5.

\subsubsection{Basis} \label{subsubsection:basis}

\emph{Spider}, \emph{fuzz}, and \emph{phaser} necessitate a choice of basis
for applying the worldly context in the conversational negation.
We can interpret this choice as determining which input density matrix sets the eigenbasis of the output, and which modifies the other's spectrum. 
We found that \emph{phaser} paired with the basis `w' (the basis of the logically negated word) performs better than the basis `c' (the basis of the worldly context) across both negations for most plausibility metrics.
This lines up with our intuition that applying worldly context updates the eigenspectrum of $\neg(\lrbracket{w_N})$, leveraging worldly knowledge to increase/decrease the weights of more/less contextually relevant values of the logical negation of~$w_N$.
However, a notable exception to this reasoning is our result that for \emph{spider} paired with $\neg_{inv}$, basis `c' has statistically significant correlations with human ratings, while basis `w' does not.

\subsubsection{Worldly context}
For these experiments, we create worldly context based on the hypernym paths provided by WordNet. As explained in Section~\ref{sec:wordnet}, we need $p_i \geq p_{i+1}$ in Equation~\ref{eq:wc_wordnet} for the more direct hypernyms to be more important than more distant hypernyms. Hence, we tried multiple monotonically decreasing functions for the weights $\{p_i\}_i$ of the hypernyms. For a word $w$ with $n$ hypernyms $h_1, ..., h_n$ ordered from closest to furthest, we define the following functions to calculate $p_i$.
\begin{align}
    \texttt{poly}_x(i) &\coloneqq (n - i)^x\\
    \texttt{exp}_x(i) &\coloneqq (1 + \frac{x}{10})^{(n-i)}\\
    \texttt{hyp}_x(i) &\coloneqq (n - i)^{\frac{x}{2}} \kE(w, h_i)
\end{align}

\begin{figure}
    \centering
    \includegraphics[width=\columnwidth]{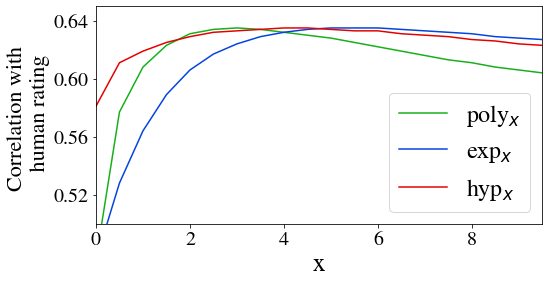}
    \caption{Correlation of results of different context functions with human rating}
    \label{fig:context_function}
\end{figure}

Figure~\ref{fig:context_function} shows on the y-axis the correlation of the human rating with the plausibility rating (\emph{trace}) of our best conversational negation (\emph{phaser} with $\neg_{sub}$ in the basis `w') and the parameters of context functions on the x-axis. We observe that all three context functions achieve a maximal correlation of 0.635, therefore being equally good. All functions eventually drop in correlation as the value of $x$ increases, showing that having the context too close to the word does not yield optimal results either. One important observation is that at $x = 0$, $\texttt{hyp}_x(i) = \kE(w, h_i)$ still performs well with a correlation of 0.581, despite not taking the WordNet hypernym distance into account. This is an evidence for the potential of the context creation based on density matrix entailment proposed in Section~\ref{sec:data-driven-context}.

\subsubsection{Plausibility rating measures}
On top of calculating the conversational negation, the experiments call for comparing the results of the conversational negation with $w_A$ to give plausibility ratings. We compare the hyponymies $\kE$, $\khyp$, and $\kBA$, as well as \emph{trace similarity}. The results show that \emph{trace similarity} and $\kEtwo$ interact most sensibly with our conversational negation, attaining 0.635 and 0.602 correlation with mean human ratings respectively. For the asymmetric measures $\kE$ and $\khyp$, computing the entailment from $w_A$ to the conversational negation of $w_N$ performed better than the other direction. For all similarity measures (except $\khypone$), \emph{$\neg_{sub}$} paired with \emph{phaser} in the basis `w' performs the best.
 
\section{Future work}
The framework presented in this paper shows promising results for conversational negation in compositional distributional semantics. Given its modular design, additional work should be done exploring more kinds of logical negations, compositions and worldly contexts, as well as situations for which certain combinations are optimal. Since creating worldly context---as presented in this paper---is a new concept in the area of DisCoCirc, it leaves the most room for further exploration. In particular, our framework does not handle how to disambiguate different meanings of the same word; for example, the worldly context of the word \emph{apple} should be different for the fruit \emph{apple} versus the technology company \emph{apple}.

Our conversational negation framework currently does not model a different kind of negation where the suggested alternative is an antonym rather than just any other word that appears in similar contexts. For instance, the sentence \emph{Alice is not happy} suggests that Alice is \emph{sad}---an antonym of \emph{happy}---rather than \emph{cheerful}, even though \emph{cheerful} might appear in similar contexts as \emph{happy}. We would like to extend the conversational negation framework to account for this.

We would like to implement the context generation method presented in Section~\ref{sec:data-driven-context} and test on the current experimental setup.\footnote{The code is available upon request.} To further validate the framework, more data sets should be collected and evaluated on to explore, for each type of relation between words, what construction of conversational negation yields sensible plausibility ratings.

For the conversational negation to be fully applicable in the context of compositional distributional semantics, further theoretical work is required to generalize the model from negation of individual nouns to negation of other grammatical classes and complex sentences.
Furthermore, we would like to analyze the interplay between conversational negation, textual context, and evolving meanings.
Lastly, the interaction of conversational negation with logical connectives and quantifiers leaves open questions to explore.

\section*{Acknowledgements}
We would like to give special thanks to Martha Lewis for the insightful conversation and for sharing her code for generating density matrices.
We appreciate the guidance of Bob Coecke in introducing us to the field of compositional distributional semantics for natural language processing.
We thank John van de Wetering for informative discussion about ordering density matrices.
We thank the anonymous reviewers for their helpful feedback.
Lia Yeh gratefully acknowledges funding from the Oxford-Basil Reeve Graduate Scholarship at Oriel College in partnership with the Clarendon Fund.

\bibliographystyle{aclnatbib}
\bibliography{RodatzShaikhYehSemSpace2021}

\begin{thebibliography}{28}
\expandafter\ifx\csname natexlab\endcsname\relax\def\natexlab#1{#1}\fi

\bibitem[{{Abramsky} and {Coecke}(2004)}]{Abramsky:2004categorical_semantics}
Samson {Abramsky} and Bob {Coecke}. 2004.
\newblock \href {https://doi.org/10.1109/LICS.2004.1319636} {A categorical
  semantics of quantum protocols}.
\newblock In \emph{Proceedings of the 19th Annual IEEE Symposium on Logic in
  Computer Science, 2004.}, pages 415--425.

\bibitem[{Ahsaee et~al.(2014)Ahsaee, Naghibzadeh, and
  Naeini}]{ahsaee:2014semantic}
Mostafa~Ghazizadeh Ahsaee, Mahmoud Naghibzadeh, and S~Ehsan~Yasrebi Naeini.
  2014.
\newblock \href
  {https://www.researchgate.net/publication/326304723_WordNet-based_Semantic_Similarity_Measures_for_Process_Model_Matching}
  {Semantic similarity assessment of words using weighted wordnet}.
\newblock \emph{International Journal of Machine Learning and Cybernetics},
  5(3):479--490.

\bibitem[{Baksalary et~al.(1989)Baksalary, Pukelsheim, and
  Styan}]{baksalary:1989properties}
Jerzy~K. Baksalary, Friedrich Pukelsheim, and George~P.H. Styan. 1989.
\newblock \href {https://doi.org/https://doi.org/10.1016/0024-3795(89)90069-4}
  {Some properties of matrix partial orderings}.
\newblock \emph{Linear Algebra and its Applications}, 119:57--85.

\bibitem[{Balkir et~al.(2016)Balkir, Sadrzadeh, and
  Coecke}]{balkir:2016entailment_using_density_matrices}
Esma Balkir, Mehrnoosh Sadrzadeh, and Bob Coecke. 2016.
\newblock \href {https://arxiv.org/abs/1506.06534} {Distributional sentence
  entailment using density matrices}.
\newblock In \emph{Topics in Theoretical Computer Science}, pages 1--22, Cham.
  Springer International Publishing.

\bibitem[{Bankova et~al.(2019)Bankova, Coecke, Lewis, and
  Marsden}]{bankova:2019graded_entailment}
Dea Bankova, Bob Coecke, Martha Lewis, and Dan Marsden. 2019.
\newblock \href {https://doi.org/10.15398/jlm.v6i2.230} {Graded hyponymy for
  compositional distributional semantics}.
\newblock \emph{Journal of Language Modelling}, 6(2):225–260.

\bibitem[{Blunsom et~al.(2013)Blunsom, Grefenstette, and
  Hermann}]{hermann:2013notnotbad}
Phil Blunsom, Edward Grefenstette, and Karl~Moritz Hermann. 2013.
\newblock \href {https://arxiv.org/abs/1306.2158} {``not not bad'' is not
  ``bad'': A distributional account of negation}.
\newblock In \emph{Proceedings of the 2013 Workshop on Continuous Vector Space
  Models and their Compositionality}.

\bibitem[{Bolt et~al.(2017)Bolt, Coecke, Genovese, Lewis, Marsden, and
  Piedeleu}]{Bolt:2017conceptual_spaces}
Joe Bolt, Bob Coecke, Fabrizio Genovese, Martha Lewis, Dan Marsden, and Robin
  Piedeleu. 2017.
\newblock \href {http://arxiv.org/abs/1703.08314} {Interacting conceptual
  spaces {I} : Grammatical composition of concepts}.
\newblock \emph{CoRR}, abs/1703.08314.

\bibitem[{Boyd-Graber et~al.(2006)Boyd-Graber, Fellbaum, Osherson, and
  Schapire}]{boyd:2006extending_wordnet}
Jordan Boyd-Graber, Christiane Fellbaum, Daniel Osherson, and Robert Schapire.
  2006.
\newblock \href
  {http://citeseerx.ist.psu.edu/viewdoc/download?doi=10.1.1.441.4494&rep=rep1&type=pdf}
  {Adding dense, weighted connections to wordnet}.
\newblock In \emph{Proceedings of the third international WordNet conference},
  pages 29--36. Citeseer.

\bibitem[{Coecke(2020)}]{coecke:2020textstructure}
Bob Coecke. 2020.
\newblock \href {http://arxiv.org/abs/1904.03478} {The mathematics of text
  structure}.

\bibitem[{Coecke et~al.(2020)Coecke, de~Felice, Meichanetzidis, and
  Toumi}]{coecke:2020QNLP}
Bob Coecke, Giovanni de~Felice, Konstantinos Meichanetzidis, and Alexis Toumi.
  2020.
\newblock \href {https://arxiv.org/abs/2012.03755} {Foundations for near-term
  quantum natural language processing}.
\newblock \emph{ArXiv}, abs/2012.03755.

\bibitem[{Coecke and Meichanetzidis(2020)}]{coecke:2020meaningupdate}
Bob Coecke and Konstantinos Meichanetzidis. 2020.
\newblock \href {https://arxiv.org/abs/2001.00862} {Meaning updating of density
  matrices}.
\newblock \emph{FLAP}, 7:745--770.

\bibitem[{Coecke et~al.(2010)Coecke, Sadrzadeh, and Clark}]{coecke:2010DMM}
Bob Coecke, Mehrnoosh Sadrzadeh, and Stephen Clark. 2010.
\newblock \href {https://arxiv.org/abs/1003.4394} {Mathematical foundations for
  a compositional distributional model of meaning}.
\newblock \emph{Lambek Festschrift Linguistic Analysis}, 36.

\bibitem[{De~las Cuevas et~al.(2020)De~las Cuevas, Klinger, Lewis, and
  Netzer}]{delascuevas:2020catsclimb}
Gemma De~las Cuevas, Andreas Klinger, Martha Lewis, and Tim Netzer. 2020.
\newblock \href {https://arxiv.org/abs/2005.14134} {Cats climb entails mammals
  move: preserving hyponymy in compositional distributional semantics}.
\newblock In \emph{Proceedings of SEMSPACE 2020}.

\bibitem[{Evans et~al.(1996)Evans, Clibbens, and Rood}]{evans:1996role}
Jonathan St~BT Evans, John Clibbens, and Benjamin Rood. 1996.
\newblock \href {https://doi.org/10.1006/jmla.1996.0022} {The role of implicit
  and explicit negation in conditional reasoning bias}.
\newblock \emph{Journal of Memory and Language}, 35(3):392--409.

\bibitem[{Fellbaum(1998)}]{wordnet}
Christiane Fellbaum. 1998.
\newblock \href {https://wordnet.princeton.edu/} {Wordnet: An electronic
  lexical database}.

\bibitem[{Grefenstette and
  Sadrzadeh(2011)}]{grefenstette:2011experimental_support}
Edward Grefenstette and Mehrnoosh Sadrzadeh. 2011.
\newblock \href {https://www.aclweb.org/anthology/D11-1129} {Experimental
  support for a categorical compositional distributional model of meaning}.
\newblock In \emph{Proceedings of the 2011 Conference on Empirical Methods in
  Natural Language Processing}, pages 1394--1404, Edinburgh, Scotland, UK.
  Association for Computational Linguistics.

\bibitem[{Hearst(1992)}]{hearst:1992automatic}
Marti~A Hearst. 1992.
\newblock Automatic acquisition of hyponyms from large text corpora.
\newblock In \emph{Coling 1992 volume 2: The 15th international conference on
  computational linguistics}.

\bibitem[{Horn(1972)}]{horn:1972negation}
Laurence Horn. 1972.
\newblock \href
  {https://www.researchgate.net/publication/247046187_On_the_Semantic_Properties_of_Logical_Operators_in_English}
  {On the semantic properties of logical operators in english}.
\newblock \emph{Unpublished Ph.D. dissertation}.

\bibitem[{Kruszewski et~al.(2016)Kruszewski, Paperno, Bernardi, and
  Baroni}]{kruszewski:2016conv_neg}
Germ{\'a}n Kruszewski, Denis Paperno, Raffaella Bernardi, and Marco Baroni.
  2016.
\newblock \href {http://dx.doi.org/10.1162/COLI_a_00262} {There is no logical
  negation here, but there are alternatives: Modeling conversational negation
  with distributional semantics}.
\newblock \emph{Computational Linguistics}, 42(4):637--660.

\bibitem[{Lewis(2019)}]{lewis:2019compositional_hyponymy}
Martha Lewis. 2019.
\newblock \href {https://doi.org/10.26615/978-954-452-056-4_075} {Compositional
  hyponymy with positive operators}.
\newblock In \emph{Proceedings of the International Conference on Recent
  Advances in Natural Language Processing (RANLP 2019)}, pages 638--647, Varna,
  Bulgaria. INCOMA Ltd.

\bibitem[{Lewis(2020)}]{lewis:2020towardslogicalnegation}
Martha Lewis. 2020.
\newblock \href {https://arxiv.org/abs/2005.04929} {Towards logical negation
  for compositional distributional semantics}.
\newblock \emph{IfCoLoG Journal of Logics and their Applications}, 7(3).

\bibitem[{Oaksford(2002)}]{oaksford:2002contrast}
Mike Oaksford. 2002.
\newblock \href {https://doi.org/10.1080/13546780143000170} {Contrast classes
  and matching bias as explanations of the effects of negation on conditional
  reasoning}.
\newblock \emph{Thinking \& Reasoning}, 8(2):135--151.

\bibitem[{Pennington et~al.(2014)Pennington, Socher, and
  Manning}]{pennington:2014glove}
Jeffrey Pennington, Richard Socher, and Christopher Manning. 2014.
\newblock \href {https://doi.org/10.3115/v1/D14-1162} {{G}lo{V}e: Global
  vectors for word representation}.
\newblock In \emph{Proceedings of the 2014 Conference on Empirical Methods in
  Natural Language Processing ({EMNLP})}, pages 1532--1543, Doha, Qatar.
  Association for Computational Linguistics.

\bibitem[{Piedeleu et~al.(2015)Piedeleu, Kartsaklis, Coecke, and
  Sadrzadeh}]{piedeleu:2015open}
Robin Piedeleu, Dimitri Kartsaklis, Bob Coecke, and Mehrnoosh Sadrzadeh. 2015.
\newblock \href {http://arxiv.org/abs/1502.00831} {Open system categorical
  quantum semantics in natural language processing}.

\bibitem[{Prado and Noveck(2006)}]{prado:2006negation}
Jérôme Prado and Ira~A. Noveck. 2006.
\newblock \href {https://doi.org/10.1080/13546780500371241} {How reaction times
  can elucidate matching effects and the processing of negation}.
\newblock \emph{Thinking and Reasoning}, 12(3).

\bibitem[{Roller et~al.(2018)Roller, Kiela, and Nickel}]{roller:2018hearst}
Stephen Roller, Douwe Kiela, and Maximilian Nickel. 2018.
\newblock Hearst patterns revisited: Automatic hypernym detection from large
  text corpora.
\newblock \emph{arXiv preprint arXiv:1806.03191}.

\bibitem[{van~de Wetering(2018)}]{vandewetering:2018ordering}
John van~de Wetering. 2018.
\newblock \href {https://doi.org/10.1063/1.5023474} {Ordering quantum states
  and channels based on positive bayesian evidence}.
\newblock \emph{Journal of Mathematical Physics}, 59(10):102201.

\bibitem[{Widdows and Peters(2003)}]{widdows:2003word}
Dominic Widdows and Stanley Peters. 2003.
\newblock \href {https://www.puttypeg.net/papers/quantum-senses.pdf} {Word
  vectors and quantum logic: Experiments with negation and disjunction}.
\newblock \emph{Mathematics of language}, 8(141-154).

\end{thebibliography}

\clearpage
\begin{appendices}
\section{Proofs} \label{sec:appendix_proofs}
\subsection{Support inverse reverses $k$-hyponymy}\label{sec:inversekhyp}
\begin{theorem}
\label{theorem:suppinvkhyp}
For two density matrices \textsf{A} and \textsf{B}, $k$-hyponymy is reversed by support inverse when $rank(\textsf{A}) = rank(\textsf{B})$:
\begin{equation}
    \textsf{A} \sqsubseteq_k \textsf{B} \Longleftrightarrow \neg_{supp}\textsf{B} \sqsubseteq_k \neg_{supp}\textsf{A}
\end{equation}
\end{theorem}
\begin{proof}
From \cite{baksalary:1989properties}, $\neg_{supp}$ reverses L\"owner order when $rank(\textsf{A}) = rank(\textsf{B})$:
\begin{equation}
    \label{eq:suppinvLowner}
    \textsf{A} \sqsubseteq \textsf{B} \Longleftrightarrow \neg_{supp}\textsf{B} \sqsubseteq \neg_{supp}\textsf{A}
\end{equation}
Thus, letting ``$\geq 0$'' denote the operator is positive:
\begin{align}
    \textsf{A} \sqsubseteq_k \textsf{B} & \Longleftrightarrow \textsf{B} - k \textsf{A} \geq 0\label{eq:inversekhyp1}\\
    & \Longleftrightarrow (k \textsf{A})^{-1} - \textsf{B}^{-1} \geq 0\label{eq:inversekhyp2}\\
    & \Longleftrightarrow \frac{1}{k} \textsf{A}^{-1} - \textsf{B}^{-1} \geq 0\\
    & \Longleftrightarrow \textsf{A}^{-1} - k \textsf{B}^{-1} \geq 0\\
    & \Longleftrightarrow \textsf{B}^{-1} \sqsubseteq_k \textsf{A}^{-1}
\end{align}
using Equations~\ref{eq:khyp}~and~\ref{eq:suppinvLowner} from Equation~\ref{eq:inversekhyp1}~to~\ref{eq:inversekhyp2}.
\end{proof}

\begin{corollary}
\label{cor:invkhyp}
For two invertible density matrices \textsf{A} and \textsf{B}, $k$-hyponymy is reversed by matrix inverse:
\begin{equation}
    \textsf{A} \sqsubseteq_k \textsf{B} \Longleftrightarrow \textsf{B}^{-1} \sqsubseteq_k \textsf{A}^{-1}
\end{equation}
\end{corollary}

\subsection{Matrix inverse reverses $\kBA$ in same basis case}\label{sec:inversekBA}
\begin{theorem}
\label{theorem:inversekBA}
For two density matrices \textsf{A} and \textsf{B} with the same eigenbasis, $\kBA$ is reversed by matrix inverse:
\begin{equation}
    \kBA(\textsf{B}^{-1}, \textsf{A}^{-1}) = \kBA(\textsf{A}, \textsf{B})
\end{equation}
\end{theorem}
\begin{proof}
\begin{align}
\kBA(\textsf{B}^{-1}, \textsf{A}^{-1}) &= \frac{\sum_i \lambda^i_{\textsf{A}^{-1}} - \lambda^i_{\textsf{B}^{-1}}}{\sum_i \abs{\lambda^i_{\textsf{A}^{-1}} - \lambda^i_{\textsf{B}^{-1}}}}\label{eq:inversekBA1}\\
&= \frac{\sum_i \frac{1}{\lambda^i_{\textsf{A}}} - \frac{1}{\lambda^i_{\textsf{B}}}}{\sum_i \abs{\frac{1}{\lambda^i_{\textsf{A}}} - \frac{1}{\lambda^i_{\textsf{B}}}}}\label{eq:inversekBA2}\\
&= \frac{\sum_i \lambda^i_{\textsf{B}} - \lambda^i_{\textsf{A}}}{\sum_i \abs{\lambda^i_{\textsf{B}} - \lambda^i_{\textsf{A}}}}\\
&= \kBA(\textsf{A}, \textsf{B})
\end{align}
using Equation~\ref{eq:suppinv} from Equation~\ref{eq:inversekBA1}~to~\ref{eq:inversekBA2}.
\end{proof}

\subsection{Composing with $\neg_{sub}$ or $\neg_{inv}$ gives maximally mixed support}
\begin{theorem}
\label{theorem:compmaxmixed}
When composing a density matrix \textsf{X} with $\neg_{supp} \textsf{X}$ via \emph{spider}, \emph{fuzz}, or \emph{phaser}, the resulting density matrix has the desired property of being a maximally mixed state on the support with zeroes on the kernel.
\end{theorem}
\begin{proof}
$\neg_{supp} \textsf{X}$ and \textsf{X} have the same eigenbasis. From Equation~\ref{eq:suppinv}, all nonzero eigenvalues of $\neg_{supp} \textsf{X}$ are multiplicative inverses of the corresponding eigenvalue of \textsf{X}.\\
We use definitions of \emph{spider}, \emph{fuzz}, and \emph{phaser} from Equations~\ref{eq:spider},~\ref{eq:fuzz},~and~\ref{eq:phaser}.  The summation indices are over eigenvectors with nonzero eigenvalue.
\begin{align}
    &\emph{spider}\textsf{(X, }\neg_{supp}\textsf{X)}\\
    &= U_s (\textsf{X} \otimes \neg_{supp}\textsf{X)} U_s^{\dagger}\\
    &= \Big(\sum_i \ket{i}\bra{ii}\Big) (\textsf{X} \otimes\neg_{supp}\textsf{X)} \Big(\sum_j \ket{jj}\bra{j}\Big)\\
    &= \sum_i \ket{i}\bra{ii} \Big(\big(\lambda \ket{i}\bra{i}\big) \otimes \big(\frac{1}{\lambda_i} \ket{i}\bra{i}\big)\Big) \ket{ii}\bra{i}\\
    &= \sum_i \ket{i}\bra{i}\\
    &= \mathbb{I}_{supp}
\end{align}

\begin{align}
    \emph{fuzz}\textsf{(X, }\neg_{supp}\textsf{X)}
    &= \sum_i x_i P_i \circ \textsf{X} \circ P_i\\
    &= \sum_i \frac{1}{\lambda_i} P_i \Big(\sum_j \lambda_i P_i\Big) P_i\\
    &= \sum_i P_i\\
    &= \mathbb{I}_{supp}
\end{align}

\begin{align}
    &\emph{phaser}\textsf{(X, }\neg_{supp}\textsf{X)}\\
    &= \Big(\sum_i x_i P_i\Big) \circ \textsf{X} \circ \Big(\sum_i x_i P_i\Big)\\
    &= \Big(\sum_i {\lambda_i}^{-\frac{1}{2}} P_i\Big) \Big(\sum_j \lambda_j P_j\Big) \Big(\sum_k {\lambda_k}^{-\frac{1}{2}} P_k\Big)\\
    &= \sum_i P_i\\
    &= \mathbb{I}_{supp}
\end{align}

\end{proof}

\begin{corollary}
\label{cor:compmaxmixed}
When composing a density matrix \textsf{X} with $\neg_{inv} \textsf{X}$ via \emph{spider}, \emph{fuzz}, or \emph{phaser}, the resulting density matrix has the desired property of being a maximally mixed state on the support with zeroes on the kernel.
\end{corollary}

\end{appendices}

\end{document}